\documentclass[]{interact}

\usepackage{epstopdf}
\usepackage[caption=false]{subfig}
\usepackage[dvipsnames]{xcolor}
\usepackage[longnamesfirst,sort]{natbib}
\bibpunct[, ]{(}{)}{;}{a}{,}{,}
\newcommand{\rd}[1]{{\color{black} #1}}

\theoremstyle{plain}

\theoremstyle{definition}

\newtheorem{benefit}{Benefit}
\theoremstyle{remark}

\begin{document}


\title{Projection: A Mechanism for Human-like Reasoning in Artificial Intelligence}

\author{
\name{F. Guerin\thanks{CONTACT F. Guerin. Email: f.guerin@surrey.ac.uk}}
\affil{Department of Computer Science,  University of Surrey, Guildford GU2 7XH, United Kingdom.}
}

\maketitle

\begin{abstract}
Artificial Intelligence systems cannot yet match human abilities to apply knowledge to situations that vary from what they have been programmed for, or trained for.
In visual object recognition, methods of inference exploiting top-down information (from a model)  have been shown to be effective for recognising entities in difficult conditions.
Here a component of this type of inference, called `projection', is shown to be a key mechanism to solve the problem of applying knowledge to varied or challenging situations, across a range of AI domains, such as vision, robotics, or language.
Finally the relevance of projection to tackling the commonsense knowledge problem is discussed.
\end{abstract}





\section{Introduction}\label{intro}






There are two major problems in Artificial Intelligence: how to apply knowledge to varied situations, and how to acquire the knowledge (learn). This paper focuses on the first. It encompasses knowledge representation and reasoning, with a focus here on (non-classical) reasoning (a second companion paper will focus on representation).
The focus is on the act of reasoning that determines if some data can be seen (or interpreted) as belonging to a particular class, not on long chains of reasoning using diverse knowledge.

A significant weakness of Artificial Intelligence (AI) systems relative to humans is the inability to apply existing knowledge to a new problem, or to a situation that varies from what they were programmed for or trained for (also called transfer ability in some contexts). 
This causes systems to fail to recognise objects or activities in new settings, or to fail to adapt skills to variations \citep{DavisMarcus2015,Ersen2017}.
The systems  have knowledge, but they are unable to apply it when the situation changes. 
This problem affects AI systems during learning phases  as well as in final deployment phases.
Deep learning, the currently dominant approach, is quite good at learning from scratch, but less able to exploit prior knowledge to learn rapidly.
 \citet{LeCunpower2018} gives the example of a machine learning to drive a car by reinforcement learning (with no prior knowledge): it might need to run off a cliff thousands of times in order to learn to avoid it, whereas most humans learn to drive a car
in about 20 hours, without crashing, which is remarkable by AI standards. Humans must be able to apply  their prior knowledge of moving bodies, momentum, gravity, friction, etc. to make their learning problem easier (see discussion in Sect.~\ref{phys}).
\citet{LakeBBSarxiv2016} give similar examples of this stark human/AI contrast for learning to play  games.
These authors point to the need for learning systems to be able to exploit models of the world.

Whether it be for learning,  or for deployed systems, there is a need to apply knowledge across varied situations.
Humans do this with ease. When training for a new task, or tackling an unexpected situation, humans apply knowledge of,  e.g., physical interactions, or social interactions, which they learnt in different contexts.
That is why humans continue to be the only viable workers in a wide range of industries where AI could not cope with the variation.
Applying knowledge to a situation is a problem of reasoning:
It is the recognition that this situation belongs to a category about which we have prior knowledge or experience, which is applicable here.
It is a very fundamental cognitive act.
Something very fundamental is clearly lacking from AI. 
This paper is about \rd{a mechanism called `projection' which plays a crucial role in the} type of (non-classical) reasoning that could get closer to human abilities, and the implications for representation.
The essence is a top-down process `projecting' conceptual knowledge onto lower level data, and thereby searching through alternative ways to represent that data. 
This requires human-like compositional models of concepts, and also iterative inference algorithms to find a good mapping (or interpretation) from data to model.

The word `projection' is used here in a way that is not synonymous with top-down processing. Instead it is a special case of top-down: it is  a particular type of top-down processing which uses a prior compositional model of a concept, and can influence the interpretation of data in order to match components required by the model (precise definition in Sec.~\ref{maths}).
There are other types of top-down processing that are not projection, because they do not rely on compositional models; for example many of the  examples of top-down processing that  are reviewed by \citet{firestone_scholl_2016} are not projection, e.g., wearing a heavy backpack makes hills look steeper; holding a wide
pole makes apertures look narrower; desired objects are seen as closer.

\rd{In contrast to classical AI, the projection approach does not assume that there is a ground truth, and that the task is to find the one `correct' mapping of predicates to the world state; instead it allows for various different mappings to be forced depending on the needs of the situation. In contrast to deep learning it does not entangle the model of a concept with the machinery which processes the data to recognise the concept; instead it has an explicit concept model which is separate from the processing machinery.} 

The idea of the mechanism of top-down `projection' is nothing new to visual perception \citep{Epshtein14298}, or computer vision (see Sec.~\ref{exist}).
However, it can be generalised  to other aspects of cognition, with potential application across areas of AI such as higher-level vision tasks, language processing, and robotics.
It is advantageous that we have already working models of projection in vision (see Sec.~\ref{vision}), this helps us to envision implementations for other cognitive tasks, and to assess the appropriateness of projection for various tasks, and expected benefits.
Matching this transition from vision to other cognitive tasks, a speculative hypothesis behind this paper is that the brain circuitry that evolved to do sophisticated interpretation in visual perception in primates may have been copied and employed to do other  cognitive tasks in humans.


The following are key contributions:
\begin{itemize}
\item Illustrating the wide applicability of one mechanism (projection) to solve problems in a range of AI application domains, at a range of levels from lower level perceptual processing to higher level abstract concepts (Secs.~\ref{video}, ~\ref{robot}, ~\ref{higher}).
\item Explaining the relevance of projection to the long-term final solution to AI, to achieve commonsense knowledge (Sec.~\ref{repr}).
\item Explaining the role of projection in making analogies (Sec.~\ref{analogy}).
\end{itemize}



\section{Projection in Humans}\label{brief}

\begin{figure}[ht!]
	\centering
	\includegraphics[height=3.4cm]{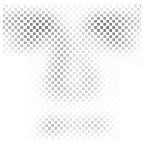}
	\phantom{kjjsdflsk}
	\includegraphics[height=3.4cm]{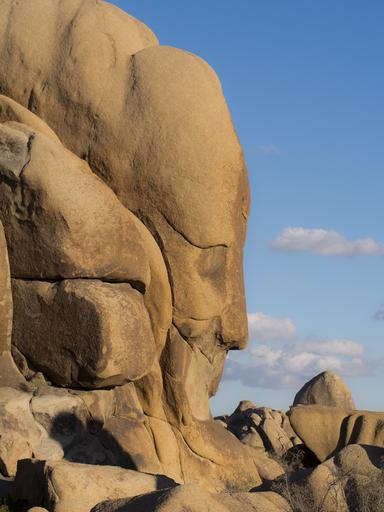}
	\phantom{kjjsdflsk}
	\includegraphics[height=3.4cm]{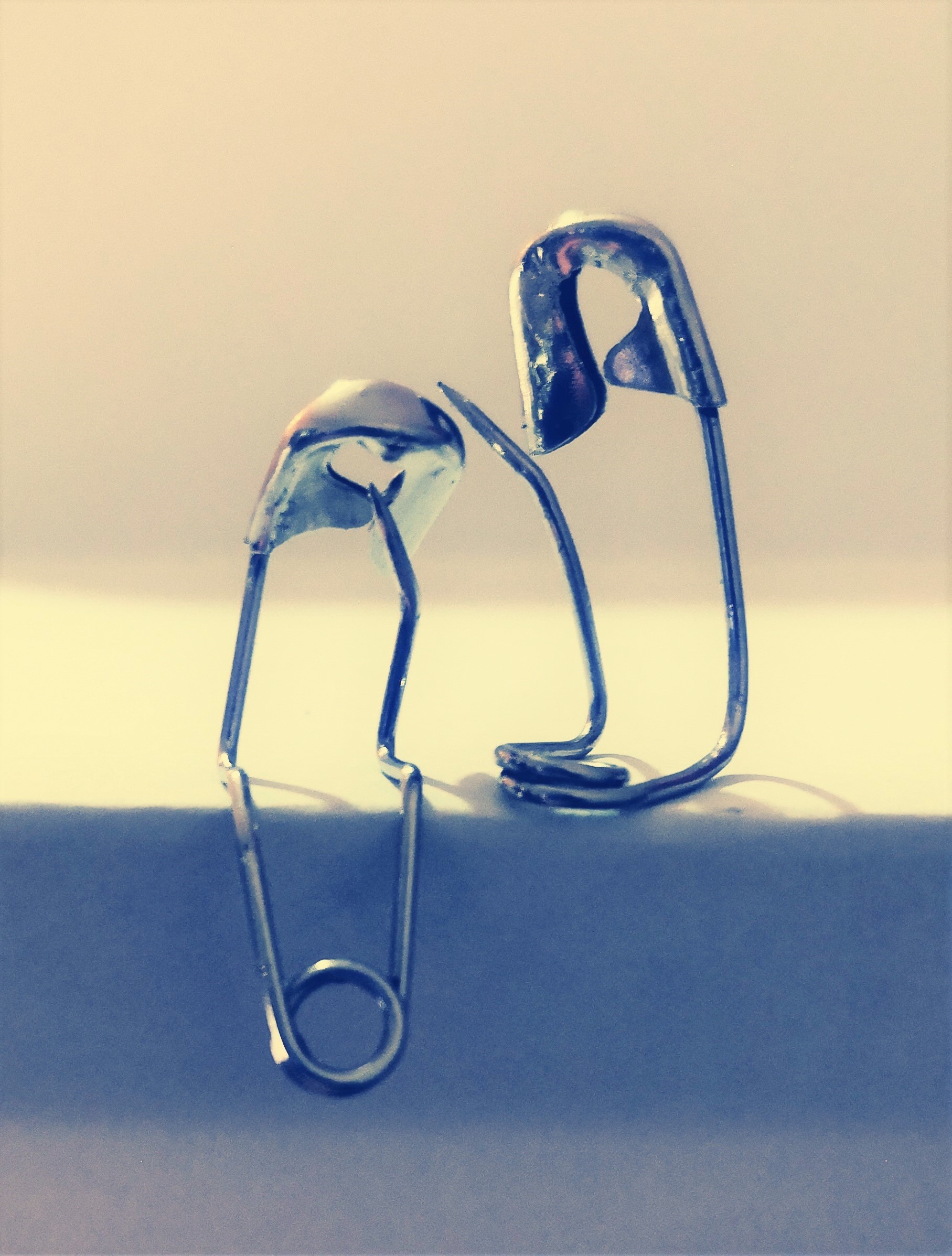}
	\caption{
	(a) Left: Low quality face image. 
	(b) Middle: What looks like a face in a cliff. (image from SnappyGoat).
	(c) Right: Advanced projection, the image could represent a human consoling another.
}\label{face}
\end{figure}

As a concrete example of projection,
consider the description of a picture, where a face structure (at a higher layer of description) can be described in terms of spatial relations among the components eye, nose, mouth, brow (at a lower layer).
The projection mechanism is a top-down process whereby a set of elements at a lower layer are grouped and interpreted as corresponding to constituent components of a higher layer.
For example in visual perception of a human face in noisy or poorly lit conditions, every low level element might not be clearly recognisable and identifiable on its own (Figure~\ref{face} (a)). When an ambiguous perceived element, e.g., arising from the mouth, is given the interpretation `mouth' partly by virtue of its relationship to other face elements, this is a result of projection.
This projection idea\footnote{We adopt the word `projection' from \citet{bipinbook}; in other texts it may be referred to as unconscious inductive
inference or inference from top-down knowledge \citep{Gregory}.}
has a long history in the psychology of visual perception, e.g. \citet{Gregory} traces it back to  von Helmholtz in the Nineteenth Century.

\begin{benefit}
Projection can assign meaning to an ambiguous piece of data by leveraging higher level prior knowledge and other available data. 
\end{benefit}

In general projection works with hierarchical compositional knowledge structures where higher layers (e.g. face) describe relationships among components in lower layers (e.g. eye, nose, mouth). Projection helps to create a mapping between elements sensed from real-world data (e.g. parts of the mouth), and the components in an abstract knowledge structure (e.g. symbol for mouth), where that structure defines the allowed relationships among components.
In perception projection works together (interactively) with bottom-up processes, to recognise objects, words, or events (\citealp{Epshtein14298}; \citealp{Samuel1981}; \citealp{INDURKHYA2006133}).

Fig.~\ref{face} (a) is made from an actual face, so  to interpret it as a face would be considered  conventional. Fig.~\ref{face} (b) is an actual cliff, but the human projection mechanism can easily interpret it as a face; this is called 
novel projection \citep{bipinbook}.
The same mechanism can produce both conventional and novel projections: in both cases
it maps an abstract knowledge structure to real-world elements. The difference is in society's consensus on what is the ground truth, or typical thing to recognise, in that real-world data; novel projection leads to recognising something not typically recognised. 
Novel projection is of interest because it is important in creativity (ibid.). Sometimes creativity is important to recognise an instance of a concept in data; it may require a new perspective to see something that a cursory analysis fails to see. 
It is worth stressing that there is no  distinction in processing (or mechanism) between conventional and novel projection\footnote{This point is proved by the fact that a computer system built for conventional projection will automatically do novel projection when presented with suitable data, as shown in Fig.~\ref{horse}.}: something might be not typically recognised simply because it is overlooked by many people, and the notion of `ground truth' becomes more open to interpretation as one moves to more abstract tasks.
What is required in either case is a way to search among alternative possible mappings between a knowledge structure and data from the world.
This type of flexible projection is important  to the AI tasks discussed in the following sections.
\begin{benefit}
Projection allows for creativity through the novel application of a concept from one domain to a different domain, or through facilitating a different perspective using concepts within a domain. 
\end{benefit}
In the example of the cliff face one may be conscious of a projection mechanism at work, e.g., on seeing eyes and nose, one may search for something like a mouth in the appropriate location. 
These difficult conditions merely make the viewer more conscious of the projection process.
In everyday cognition most projection happens subconsciously. 
Projection must be happening all the time in human perception, because typical viewing conditions are difficult,  due to the many `defects' in the human visual apparatus, especially in the periphery of vision \citep{oregan1992}.
For example, blood vessels and nerve axons obscure significant areas of the retina, and there is sparsity and nonuniformity in the distribution of cones.
\begin{benefit}
Projection assists in the recognition of concepts in difficult conditions. 
\end{benefit}

The more advanced example of projection in Fig.~\ref{face} (c) goes beyond purely visual concepts as it involves affective states and social interaction. Understanding this image requires an iterative inference where bottom-up vision detects general form, triggering the possibility that this represents the human form; this in turn triggers a projection to imagine the human form in a matching pose, leading to the recognition of the affective states and the social interaction. The artist who originally created the idea of this composition performed an analogy (Sec~\ref{analogy}) where the relationship among parts in the human domain is projected onto the available objects and parts in the safety-pin domain. The higher-level relations in the human domain (e.g. bending of the head or arm) organise lower level components in the safety-pin domain to form the same relationship. There is no actual arm in the safety-pin, but when the relations from the human domain are imposed on the parts of a safety-pin then we can recognise the part in the appropriate relationship as an `arm'. To perform human-like projection in machines, the machines will need to have human-like part-based compositional models of concepts.

\rd{
Projection is also required to generate images such as Fig.~\ref{face} (c).
There is currently some interest in using deep learning to generate artistic works which showcase an artificial `imagination', e.g. with Deepdream \citep{dream2,DeepDream}. There are also efforts to encourage creativity by forcing deviation from established styles (Creative Adversarial Networks \citep{DBLP:journals/corr/ElgammalLEM17}).
However,  current work does seem to be limited by its lack of a crisp symbolic compositional model of concepts. Despite significant recent work in the generation of artistic images, the state of the art does not seem to be capable of producing something like Fig.~\ref{face} (c).
When a deep learning system is tasked with imposing the appearance of one concept on another it tends to introduce spurious artefacts and distortions in the target concept, which e.g., in the case of the paperclip, make it no longer look like an actual paperclip that has been bent.
Deep learning systems do learn something about the composition of objects, but it is in a messy entangled way, not clearly differentiating necessary from incidental features in training examples, which leads to artefacts in generation.
Visual analogies require an understanding of objects and their parts, ideally in the form of crisp compositional models, to be able to `project' the structure of one object onto the parts of another.
Such top-down direction would seem to be required to produce visual analogies \citep{icaart09}, such as appear in surrealist artwork or advertising. 
}

\subsection{Projection beyond Visual Perception}
The above examples are in visual perception (at least initially), which is a special case of more general recognition or judgement. Consider the non-visual example of sitting on a cushioned seat and then recognising that it is wet. The immediate sensation of cold may trigger the wetness concept and, top-down, an expectation of stickiness of clothing when one moves. (The human body has no sensors specifically for wetness; this concept has a complex mapping to sensors for touch and cold and their relation with skin movement.)
In higher level cognition, recognition can be called judgement, e.g. in reading a text or watching a movie, and making a judgement about whether or not a character is wealthy. The low level elements become details of behaviour or social interactions, and the higher level concept is an elaborate model of what it means to be wealthy.
Recognition (or judgement) is \rd{a} fundamental act of reasoning; i.e. deciding that something belongs to a class or category.
Here are further examples that span varied domains:
judging if a food item (e.g. pancake or omelette) is sufficiently similar to a cloth (with similar `foldable' properties) such that  a similar manipulation can be transferred; 
 looking out the window and deciding if the utterance `it is raining' is true \citep{bipin2016}; 
judging if a sequence of transactions in a company's accounts indicates that an insider fraud is taking place, or more generally, assessing a set of facts and judging if a crime is likely to have been committed by a person. 

\rd{
\subsection{Projection in Planning}
Consider the type of thinking involved in planning physical manipulation activities, for example rearranging objects in a kitchen cupboard. There will be a main goal state, involving desired positions of main objects, and a series of steps will be planned. At each stage there are several potential candidate sub activities to consider, such as temporarily placing one item on others, or pushing one item aside. In each of these subactivities projection is employed, e.g. to make the judgement about whether or not  an object can support another, or if the upper one will roll or fall off. The objects in question must be modeled by some mental representations, e.g. of exemplar concepts (cylinder, flexible bag, etc.). Objects are approximated by mental representations, and this is a process of projection that imposes some model on an object. Projection is also used in the selection of manipulation actions to employ: given a need in a step (e.g. to open a gap between objects for inserting another) the current situation can be matched to similar remembered situations where a similar problem was solved. Projection is used to match a stored model of a remembered situation to the current situation. Projection can again be employed to foresee the consequences of steps: remembered episodes (including effects) can be projected onto the current situation.
\citet{forbus1997qualitative} describe how  analogy can be employed to find similar remembered behaviours, and to use these for mental simulation. 
In this way projection is a component mechanism that works as part of a larger machinery of cognition, in order to complete physical cognition tasks.

The same ideas can also extend from physical manipulation to  more abstract domains of reasoning, for example reasoning about a criminal case. Motives can be projected onto actors, given observed or conjectured events. Likely plans of actors can be conjectured, with projection employed at each step, e.g. to match the situation under consideration to familiar episodes one has reasoned about before.
}

\subsection{Summary}
In all cases above the bottom-up data in the lower layer is ambiguous; projection will assign an interpretation to a set of lower layer elements which supports the recognition of the higher layer concept, if sufficient evidence is deemed to be present. The mechanism of projection is precisely what \citet{hofstadter2001analogy} calls  high-level perception, and many more excellent examples are given in \citet{hofstadter2013surfaces}.
In summary,  \rd{projection is an important component mechanism in human cognition at several levels and across domains}\footnote{Indurkhya's projection \citep{bipinbook,INDURKHYA200716} is synonymous with Jean Piaget's assimilation in many contexts.}, in the remainder we see how it applies to AI.

%

\section{Projection in Computers}

\begin{figure}[ht!]
	\centering
	\includegraphics[width=\textwidth]{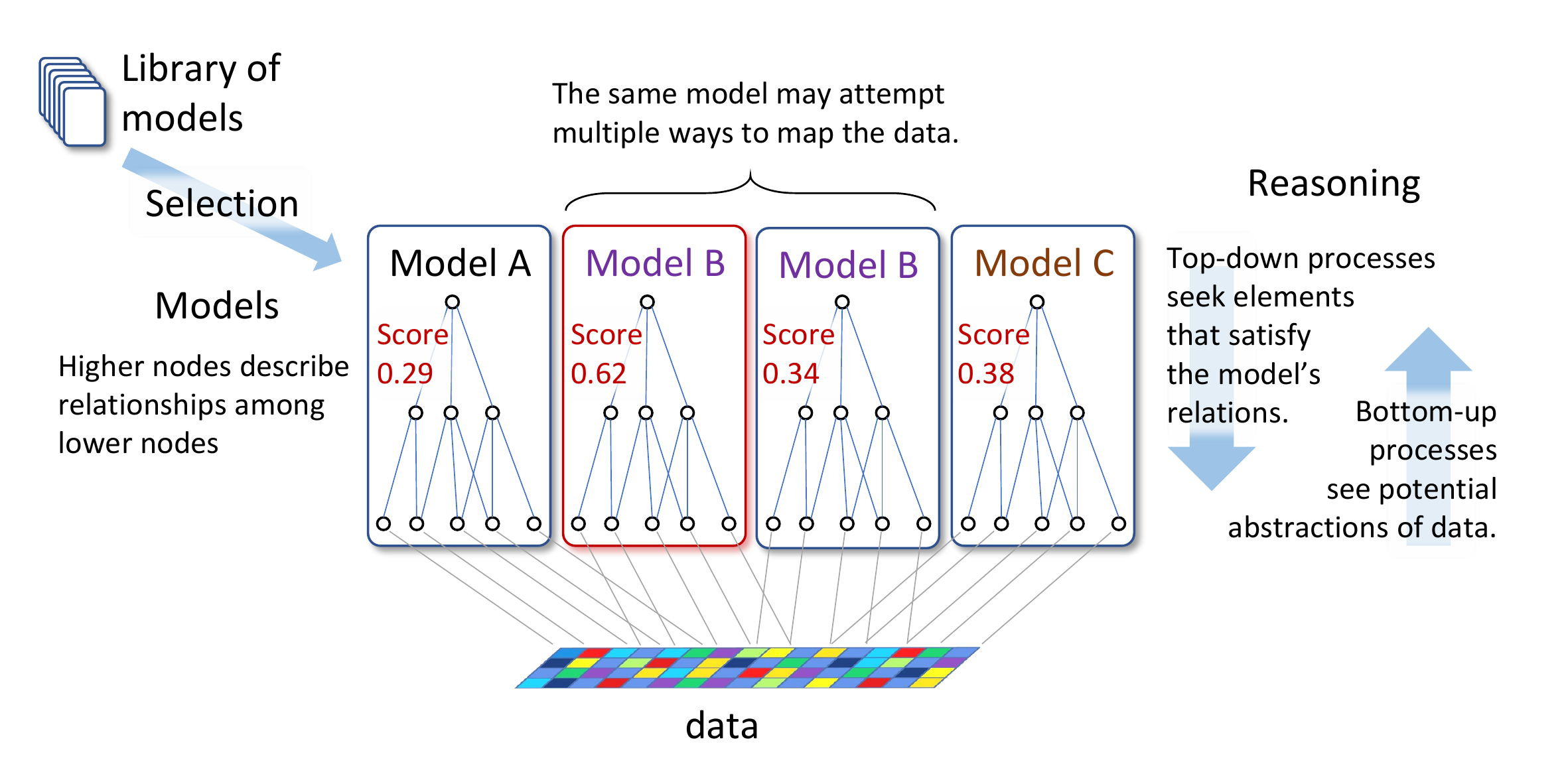}
		\caption{
	Conceptual overview of iterative reasoning where bottom-up processes interact with top-down projection to find the best interpretation of data (where an interpretation is a  mapping of the elements of a model to the data).
}\label{project}
\end{figure}

Projection can be implemented in various ways, as shown in the next section. Figure~\ref{project} illustrates the general idea, to introduce the terminology for subsequent sections.
The \textit{model} is the main data structure employed, and is structured hierarchically, with a minimum of two levels, the higher levels describe relations among  lower level elements; as in `face' describing relation among `mouth', `eyes', etc., in the previous section. Multiple models are stored in  long term memory, and selected as candidates for interpreting data. The selection process is beyond our scope, but could use task context, or other contextual factors making a particular interpretation likely. Selected models become instantiated in a workspace for the \textit{reasoning} process. This involves an interaction of top-down and bottom-up processes for determining how well a particular model could fit the data. The particular assignment of elements of a model to elements of data is called a \textit{mapping} or interpretation. There may be several instances of the same model mapping to the data in different ways, e.g. seeing a face in rock with different mappings to elements.
The reasoning process also coordinates across the various mapping attempts by seeking  the best set of mappings that could coherently explain the data (i.e. `explaining away'), or enforcing a mapping that satisfies some other goal or external pressure, e.g., if in manipulation we want to see a particular affordance, or if we are asked to apply a particular model. Benefit 1 of projection above (disambiguating data) can be explained in other words as follows: projection is when a low-level element of data could equally well be interpreted as an $x$ or a $y$ based on the local data, but I decide to call it a $y$ because that helps me to have consistent evidence for the recognition of a higher level concept, with $y$ as a constituent part. I am forcing the interpretation on the data, or projecting. In the case of novel projection $y$ is not a $y$ at all, and I know that, but treating it as a $y$ helps me to complete an analogy that may have some advantage; e.g. if a robot treats the broad base of a frying pan as like the head of a hammer and can then use it to hammer something.

The dominant approach in much of recent AI is a purely bottom-up process to go from low level data to a judgement about a concept. This pathway needs to be broken open to insert search processes in the middle which can search among possible alternative mappings between the concept (and its subcomponents) and the lower level data. Each possible candidate mapping that is searched entails a different reorganisation of the low level data. 
Simple implementations of projection can generate candidate mappings with a search process, and use a trained scoring function to evaluate candidates \citep{BENYOSEF201865,DBLP:journals/corr/abs-1710-04970}; here the top-down signal is implemented by the scoring function's selection. More complex implementations can use more message passing between layers to `agree' on a suitable mapping \citep{Georgeeaag2612}. 


\rd{
\section{Mathematical Definition}\label{maths}
A model $M$ (corresponding to a concept) is an n-tuple of levels $( l_1, l_2\ldots l_n )$ where each $l_i$ is an m-tuple $( e_1, e_2\ldots e_m )$ of elements $e_j$. 
In level $l_0$ elements are empty slots in the abstract model, and will be filled with data during interpretation. 
For levels $i> 1$ each element in level $l_i$ needs to describe a relation among some elements in level $l_{i-1}$; this is achieved by each element $e_j$ being a pair $( p,f )$ where $p$ is a tuple of specific `parts', i.e. elements from the lower level, and $f$ is a scoring function mapping parts $p$ to a real value. The scoring function $f$ assigns a  value to $p$ according to how well it captures the intended relation among parts. Since each part can itself be composed of subparts all the way down to the lowest level, the function $f$ has access to the full information of the hierarchy of parts below it. Level $l_n$ (top of hierarchy) always has only one element, ensuring that there can be a single overall score for an interpretation by a model.

$D$ is a set of data points (which could be e.g. $( x, y, \mathit{colour} )$ values of pixels).
An interpretation or mapping of $D$ by model $M$ is a binary relation over  $D$ and $l_1$ where  elements of $l_1$ are each related to at most one (not necessarily unique) element of $D$. Some elements of $D$ and $l_1$ may be unmapped (corresponding to, e.g., missing, irrelevant or unclear data). If we populate $l_1$
with the data elements mapped by the interpretation, then we can apply the scoring function of $l_n$, and determine how good the interpretation is.
Finding a good interpretation can be called inference or reasoning, and can be complex. A simple way is to run the $f$ functions of level $l_2$ over multiple subsets of $D$ to find good candidates for $l_2$ elements, then run $l_3$'s $f$ functions over these candidates to find good candidates for $l_4$, etc. This is roughly how current feedforward networks work.
This may however miss out on a good interpretation because the lower level scoring functions lack information about what is needed at the top level; it may be that some data points that produce a poor score for a certain $l_2$ element should nevertheless be used because the overall model fit will be good.
Therefore a superior inference strategy uses top-down information to guide the selection of parts in the lower levels. We call any interpretation using top-down information `projection'; i.e. where the $f$ functions from a level higher than $l_i$ contribute to the decision about what elements from $l_{i-1}$ to use.

This definition of projection is quite general, such that feedforward neural networks with skip connections can implement projection (see Sec.~\ref{DNN}). However more human-like projection requires that elements in levels are close to what humans appear to be using (i.e. the model is more human-like), with parts that are meaningful to humans, as in the work of \citet{BENYOSEF201865} for example.
}
\section{AI Applications}
This section explains how projection has been applied in vision, and how it could potentially be applied to other popular AI tasks. It also highlights the serious difficulties for existing techniques in these tasks, and explains how projection offers a way to overcome these.

\subsection{Projection in Computer Vision}\label{vision}

Sec.~\ref{exist} gives concrete examples of how projection has been implemented computationally for visual recognition tasks. Sec.~\ref{video} explains how it could be applied in higher-level visual tasks.

\subsubsection{Existing Implementations of Projection in Object or Character Recognition}\label{exist}

\begin{figure}[ht!]
	\centering
	\includegraphics[width=12cm]{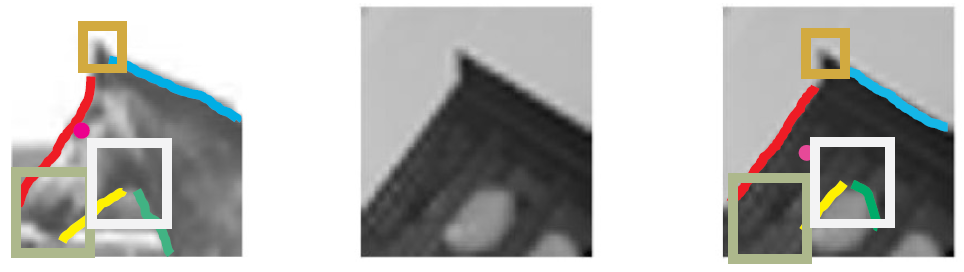}
	\caption{
	\citeauthor{BENYOSEF201865}'s (\citeyear{BENYOSEF201865}) full interpretation system can perform novel projection, interpreting a man-made object as a horse's head. Left: the model's contours, point and regions (rectangles) mapped to a photo of an actual horse's head (conventional projection/interpretation). Middle: the man-made object. Right: The mapping of the horse contours, point and regions to the man-made object (novel projection/interpretation). These images are reprinted from \citet{BENYOSEF201865} with permission from the author. 
}\label{horse}
\end{figure}

\citet{BENYOSEF201865} present a 
two-stage process which firstly works in a bottom-up feed-forward manner to identify one or several objects classes for an image, and secondly projects previously learned models of those detected classes top-down, to confirm if that object class is indeed present and give a detailed interpretation if it is\footnote{The authors do not explicitly call it projection, although at a key part in the Appendix, describing the mapping from model to image, they do say ``We project ground truth annotated contours on an edge map''.}. Their model of an object class (e.g. horse's head, as in Fig.~\ref{horse}) consists of primitives and relations among primitives. Primitives consist of contours, points, and regions. Relations include, e.g. for contour-contour relations: relative location of contour endings, smooth continuation between contour endings, length ratio between two contours, and contour parallelism.
 
 The projection occurs in the top-down process forming a detailed interpretation for a given novel image of suspected class $a$. The relations and primitives described in the model direct a search for primitives and relations in the image which could match what the model requires.
 The only relations tested for the primitives are the relations stored in the model of class $a$. 
 The result of the interpretation process is that the system has a mapping from detected primitives in the image  to known primitive parts and relations in the model for that class, and so could potentially explain its interpretation in a human-like way; e.g. this is the top contour of the horse's head, this is the bottom, this is the ear, etc.
 The system has a higher accuracy at recognising small image patches than feedforward CNNs.
 \rd{The average precision (AP) for a CNN was 0.78, while the model using interpretation had AP 0.87. More significant than this performance difference however was the sudden drop in performance of the interpretation model when the image patch is cropped further; this matched the type of drop in performance for humans trying to recognise the patches, whereas the CNN model degrades more gradually.}
 A side-effect of this human-inspired approach is that  the models learned for objects, in terms of contours etc., can be used for novel projection. For example, Fig.~\ref{horse} shows what happened  when the model learned for horse's head  was asked to produce an interpretation for a man-made object. (The authors worked to modify the model to reject such false positives, but the process can easily be relaxed to encourage such visual analogies.)

The work was  subsequently  extended  to tackle agent-agent
interactions such as `hugging', `shaking hands', `consolation' or `helping' \citep{BenYosef2018ImageIA}. This involves the model capturing  relations between elements in the two different agents, i.e. moving beyond the relations within a single object/agent.
The authors also note the further extension of image understanding ideas to more abstract concepts:
\begin{quote}
The term `image understanding' as used here depends on the image
interpretation process, but it can be more abstract, in the
sense of using concepts which go beyond components of
the physical world and relations between them, for example,
goals, moods, judgments such as ‘dangerous’ and others. \citep{BenYosef2018ImageIA}
\end{quote}

A second example of projection appears in a system recognising the distorted texts of CAPTCHAs \citep{Georgeeaag2612}. Here the models are hierarchical compositional models of text characters; e.g. `\textsf{A}' being composed of various oriented line segments and angles.
The reasoning employs a `recursive cortical network' (RCN) which has a hierarchical compositional structure, and performs a forward pass (bottom-up) to generate initial hypotheses about the letters present,  followed by a backward pass (top-down, corresponding to projection) to test the hypotheses and explain the data
, performing inference by Belief Propagation. 
For example, if the letters \textsf{A} and \textsf{K} are partly overlapped in an image, a letter \textsf{V} could be hallucinated, and is one of the suggestions (in addition to \textsf{A}, \textsf{K}) after the bottom-up pass. However the \textsf{A} and \textsf{K} can explain the image without need for the \textsf{V}, so it can be rejected.
The RCN also has lateral connections which coordinate the choice of features grouped to form a higher-level feature; this ensures smooth contours in letter shapes (actually implementing some gestalt principles). 

\rd{
The RCN CAPTCHA recogniser of \citet{Georgeeaag2612} was massively more data efficient than competing CNN approaches which required 50,000-fold larger training sets. In addition the RCN was much more robust to variations in input (e.g. spacing of characters, length of strings), whereas the CNN performance degraded dramatically.  
}

The two implementations of projection reviewed above are quite different, but both bring significant benefits to their respective tasks, when compared with feedforward models. Other projection examples in vision also exist \citep{Epshtein14298,murray2005,leonardis2008,MALIK20164}
. It suggests that projection is quite a general principle  for recognition, so that a variety of different implementations of the general principle can bring advantages. Nevertheless, examples of top-down processes are in general rare in current mainstream computer vision, which is  mostly purely feedforward.

\subsubsection{Applying  Projection to Recognising Activity from Video}\label{video}
Firstly consider short-term, or `atomic' actions: 
actions that are particularly difficult in the ActivityNet Version 1.3 dataset include smoke, grab and hug; these actions  require fine-grained discrimination \citep{DBLP:journals/corr/GuSVPRTLRSSM17}, because many of these actions can only be discriminated from other similar actions by a detailed analysis of the relations among object parts (explored in detail for still images by \citet{BenYosef2018ImageIA}). Indeed hug is one of the primary  actions tackled by \citet{BenYosef2018ImageIA}  with their model-based approach (using projection) on still images, which can analyse parts of the people in detail, e.g. the hand of one person and the back of another, and their relationship in the image. This fine-grained analysis allows the system to discriminate between hugging and superficially similar actions such as fighting.\footnote{\citet{BenYosef2018ImageIA} compared their approach against a  very deep convolutional neural network model trained on the same examples as their own approach, and  showed significantly better precision for their approach (0.80 vs. 0.69).}
This shows that in video analysis we could improve recognition by applying  projection to still images within the video. 

Moving to temporally extended activities, one can extrapolate (although not tried yet) to expect that applying the projection idea across temporal sequences would boost the performance for temporally extended activities; i.e. the primitive elements would be interactions of objects and agents at a certain time point, and the higher order relations would be temporal relations among primitive elements, possibly extending to further higher layers in more abstract activities. 
For example, a model of loading goods into a vehicle can begin with a person carrying something (recognisable by characteristic posture and gait), followed by contact with the vehicle, the loading activity, and then the unencumbered person departing. Most of the same atomic actions in a different order would be a model of unloading.
The atomic actions on their own are ambiguous and can be given meaning by the overall activity model projected; especially if working in noisy conditions where atomic detections are only partially accurate.
The projection here across the temporal domain shares the characteristics that make projection appropriate in the spatial domain for still images.
Ambiguity of activity components is common in activities that existing video classification techniques struggle with, for example, the motion of the hand descending from the mouth  could be part of  smoking or part of eating.

Disambiguating the activity in a video is  similar to the segmentation problem in CAPTCHA recognition, where it is not clear what character a certain fragment of image belongs to;  by activating several possible candidate characters, and projecting them back down to identify what other parts of the image could be part of that character, we can come up with a plausible explanation for the data. The analogous situation in video is where a temporal segment of video could be part of several longer-term activities (but not all of those are actually happening in the video). By activating models of the possible longer-term activities and projecting them down to the data one could find the activities that best explain the data, and reject the possibilities that are not required to explain the data.

Another analogy between recognising events across time and recognising CAPTCHA characters is that events can be distorted, e.g. household activities, by being interrupted by part of another activity, or being done in a non-standard way with non-typical motions. Again, the top-down fitting of models can accommodate for this by allowing for deformation of where parts of activities occur, just like parts of characters in a CAPTCHA.
Furthermore there is the complexity of relations that needs to be computed, which suggests the need for top-down direction to allow the system to be selective about what relations must be computed. For example, to discriminate between difficult classes in ActivityNet \citep{7298698} such as `violin playing' and `inspecting a violin' one might need to compute relative pose between human and violin. Complex relations are also required to discriminate mixing drinks from making a cake.
All of this requires models of the activities in time, something which can potentially be learned from data \citep[see e.g. ][]{Dubba:2015:LRE:2831071.2831073}.

The state-of the art performance in video activity recognition is much worse than in object recognition, hindering multiple real-world applications; hence  there is a strong need for new approaches.
In a 2018 (AVA Dataset) competition to classify human actions and localise them (in time and space in the video) the best mean average precision\footnote{average over all videos, mean over all classes} was 21\% \citep{DBLP:journals/corr/abs-1808-03766}. The classification turns out to be the main difficulty in this challenge \citep{DBLP:journals/corr/GuSVPRTLRSSM17}.
The authors of the AVA challenge hypothesise that better recognition will require rich models of motion and interactions \citep{DBLP:journals/corr/GuSVPRTLRSSM17}. 
The leading approaches did not use top-down processing \citep{tsinghua2018,DBLP:journals/corr/abs-1807-10066}.

Further insights about what makes activity recognition difficult are given in the 2015 ActivityNet challenge \citep{7298698}, which is not restricted to atomic actions. Firstly they note that  activities spanning a longer time are the hardest to classify\footnote{In recognising activity from video there is a range of increasingly abstract tasks (and increasing difficulty), forming a hierarchy from simpler atomic actions through to more abstract temporally extended activities, as explained by \citet[citing Barker and Wright]{DBLP:journals/corr/GuSVPRTLRSSM17}; e.g.  let us use `$<$' as a symbol for `is a sub-element of' (at lower layer):  `stepping down from the curb'$<$`crossing street'$<$`walking to school'$<$`working to pass from the third grade'$<$`getting an education'$<$`climbing to the top in life'. The atomic actions such as `stepping down from the curb' require a relatively simple analysis of objects (human, curb) and their relationship in space and time, whereas the more abstract end of the hierarchy would require extremely elaborate models of the concepts involved, and a video analysis spanning a long temporal duration (e.g. in a movie).}, this is not surprising given that the typical techniques used do not employ temporally extended models of activities. Secondly they note that similar activities cause confusion, especially with similar objects or contexts, and similar motions; this again is not surprising because the techniques use features derived from networks for object recognition, with some short range temporal information, and motion features, but no long-range temporal models.
Household activities (e.g. sorting laundry, mowing the lawn, recycling) were the hardest category of activities. The structure in these activities is highly variable at the low level, e.g. activities might be interrupted and resumed, and might happen in varying orders, but there still is structure, it just requires a more sophisticated model to account for the possible variation. This is in stark contrast to activities in the category sports and exercise which ``typically contain a repetitive and structured temporal
sequence and are usually performed in similar scene contexts'' (ibid.), existing feedforward models work very well on these.
Overall,  temporal action localization (i.e. determining when a particular action starts and ends in the video) ``remains as one of the most challenging unsolved problems in computer vision'' \citep{DBLP:journals/corr/abs-1808-03766}.

\subsubsection{Applying  Projection to Higher-Level Vision Tasks}\label{higher}

Higher-level tasks employ more abstract concepts; e.g. in vision:  image captioning\footnote{Current image captioning tends to perform very badly on out-of-training-distribution images, see e.g. twitter.com/INTERESTING\_JPG. Some kind of reasoning seems necessary.} \citep{Anderson_2018_CVPR}, answering questions about images (visual question answering) \citep{DBLP:journals/corr/KafleK17}, and recognising more abstract concepts in   videos  \citep{semanticgap}. These  can require recognising  abstract relationships among entities (e.g. friendship), which are more challenging than object recognition due to the variety of surface appearances (which entails the need for more training examples for feedforward neural network training). 
A model-based approach using projection can map a model to a wider variety of surface realisations without needing to see so many examples, as demonstrated by the results of \citet{Georgeeaag2612}. 
Projection requires a model of the concept in question. For example, for `friendship' it would explicitly describe observable relationships among people such as participating in an activity together, eyegaze, smiling, etc. Acquiring such models is a challenge yet to be tackled.




A further vision task where projection is required is visual context. It is surmised that there are `context frames' which are structures encoding knowledge about likely objects in a particular scene and their relationships \citep{bar2004};
  a context frame becomes activated by rapid coarse processing of a scene, and thereafter  exerts top-down influence on the perception of elements of a scene (ibid.). For example, \citet{Torralba2003} has shown how the same shape in a blurred outdoor street scene could be perceived as a car or as a pedestrian depending on its position and orientation. This suggests that the context frame for a street scene is a probabilistic model which allows for buildings, streets, vehicles, and pedestrians, only at certain locations, scales, and orientations, relative to each other. Given a blurred image with recognisable building to the rear, and street in front, a horizontally elongated blurred object on the ground at the far end of the street is most likely a car, but the same blurred object rotated 90 degrees, and on the street nearer the observer, is most likely a pedestrian. According to this formulation the type of model for a face described in Sec.~\ref{brief} may well just be a special case of the more general class of models that could include a context frame.




\subsection{Projection in Robotics}\label{robot}
A projection mechanism could help robotics to advance in applications in open environments (e.g., everyday tasks in the home or a workplace with varied objects or scenarios) where current performance is very far short of human level \citep{Ersen2017}. While robots can be programmed for specific tasks with specific materials, they do not generalise their behaviour well when there are slight changes in conditions. See for example the struggles to pick up products in the Amazon Picking Challenge \citep{DBLP:journals/corr/CorrellBBBCHORR16}. Similarly for household tasks; while there are robots programmed to fold clothing \citep{MillerBFDGA12}, that will work for certain clothing, it is quite a leap to consider how a robot could transfer any of that knowledge to the case of a folded pancake \citep{pancakes11humanoids}, which requires specific programming. Humans however, even young children, easily make transfers across domains. The ability to transfer in a human-like way would go a long way to solving robotics challenges in open environments. Such transfer is surmised to require an element of creativity, working by creating a mapping between scenarios \citep{fitz2017}. This is the mechanism of analogy, which is hypothesised to  require a top-down process that reorganises bottom-up perceptual data  \citep{DBLP:bibsonomy_hofstadter95,bipinbook}. See further discussion in Sec.~\ref{analogy}.

The remainder of this section speculates on how projection could be applied in robotics, where models are robot skills combining perception and action (i.e. where the motions involved are parameterised by values returned from perception; see e.g. \citet{PawelICRA}). 
Projection can allow a skill to be transferred across varied situations. Ideally a robot should be able to act similarly to a human toddler who learns a new manipulation skill and then is eager to apply it to every situation that affords the opportunity (much like the saying that ``If all you have is a hammer, everything looks like a nail.'').
Consider for example a ``peg in hole'' skill which includes perception routines to detect a hole, and its orientation, and to recognise the shape and major axis of the (cylindrical) peg. Assume the robot has been coded with a skill that consists of a first phase where the end of the peg is pressed to the aperture, but not in the correct orientation for insertion, followed by a second phase where the peg is kept in contact with the inner edges of the hole (using force feedback) while being rotated into the correct orientation, followed by a final phase where it is pushed into the hole. Each of the phases of motion will be parameterised according to the keypoints (e.g. nearest edge of aperture) and orientations extracted by perception for a particular scenario. To enable projection the perception routines need to use a hierarchical compositional model: the aperture of the hole must be modeled as consisting of relationships among neighbouring edge segments, surrounding a space (similar to how \citet{Georgeeaag2612}, described above, would model the letter `o'). The inner walls of the hole must be likewise modeled as a composition of wall segments. This model can then be projected on varied scenes to identify approximate holes, for example  in a crate packed with  groceries where there is a gap  creating an approximate hole; here the items surrounding the gap have enough edge components in a suitable configuration to approximate an aperture. When the aperture and hole orientation are identified the motor skill can be applied to insert e.g. a bottle. The skill can likewise be generalised to an open jar of food for inserting a cooking utensil. To deal with  complex varied shapes we could first approximate shapes with a coarse mesh, before searching for elements of an aperture.

In this projection the perceptual process is driven top-down. Starting with a need to achieve a task, this triggers likely skills, and then the perceptual elements that are prerequisites of those skills are selected for. E.g., to attempt to see possible holes, even if only approximate matches. Perceptual grouping processes (e.g. seeing separated edges or surfaces as one long edge or surface) and perceptual approximations (e.g. ignoring fine grained detail, or approximately fitting geometric shapes) can be influenced by top-down pressure. This top-down direction overcomes the problem of too many potential approximate matches for perceptual abstractions if they were all triggered bottom-up from the environment. When the task exerts a top-down pressure in this way we can say that the perception or the representation is `task-driven' \citep{8731700,ZHU2020310}. A disposable representation can be created as needed at a specific time, and a different representation can later be created for the same scene, if the task is different.		This also exposes the forcing aspect of projection: in mapping a model to a situation it can create similarity  rather than simply record an existing similarity \citep{kittay1982creation}.

A second example shows how the perceptual elements matched do not need to all belong to one feature such as an aperture, but can be different parts referenced in the skill.
Consider a group of rectangular cereal boxes in a shelf, tightly packed  like books in a bookcase, so that the  smaller side surface of each box is facing the robot, and each box is wedged between others. Assume the robot has a skill for removing a box by first exerting pressure on the top surface and pulling back towards itself, causing a rotation of the box, and exposing the left and right sides for a subsequent grasp.
Let us call this the skill ``pull surface to rotate''.
The key environmental elements in this model (skill description) are the top surface which will be pushed, the surface of the shelf on which the box stands, which resists the push, and the direction of the pull, where there must be free space.
When the robot wants to transfer the skill to a   different scenario it needs to make a mapping between the key model elements and entities in the new scenario.
This is effectively assigning an interpretation to the elements of the new scenario as parts of the skill description (similarly to how contours in an image could be assigned an interpretation as parts of a face or horse's head above).
For example: a pizza box lies flat in a freezer. The robot needs to carry out a search through possible mappings from its skill elements to the scenario elements. 
In this it searches through the available surfaces of the new target object, and the directions in which the motion could be applied, and simulates the effects of these, eventually choosing to push on a side surface and rotate the box upwards. This search process requires an interaction between bottom-up information from what the environment affords, and the top-down needs of the task dictating the elements to search for, together with simulation to try out potential mappings of the skill to environmental elements. An implementation of this does not exist. The general idea is discussed further in Sec.~\ref{inference}. 
\begin{benefit}
Projection can allow skills to be transferred to novel scenarios.
\end{benefit}
From the perspective of the projection process this is really no different than what has been called Benefit 2 or 3 above, but for robotic skills some new processes need to be added to check the physical plausibility of a proposed mapping from a scenario to a skill (which were not required in e.g. recognising a face in a cliff).
The use of projection to transfer skills can be thought of like the toddler applying a new skill to everything; one should be able to attempt a mapping from a skill to any scenario, e.g. to push on a surface and pull to rotate an object. In many scenarios the result is poor, much like one can try to see a face in any cliff, or in a pattern of clouds. This is also the idea behind doing `the same thing' in `Tabletop' \citep{DBLP:bibsonomy_hofstadter95}. 
In the other direction one could fix the scenario and apply all skills in the library to see what is produced.
This is a productive aspect to projection: an agent can generate multiple different possibly novel representations of a scene by applying its skills. Some of these may lead to interesting creative solutions to a manipulation problem.
It is also worth appreciating the creativity in mundane everyday manipulations.  Everyday activities in unstructured environments involve the interpretation of scenarios that were never encountered in exactly the same configuration before. Also the skills applied are typically unique (in their details) to the agent that has learned them. That is why the act of projection can be seen as one ``through which a cognitive agent asserts its
creative spirit onto the environment'' \citep{INDURKHYA2006133}.

The above description of the pull-to-rotate skill omits the steps involved in progressing from learning a skill (e.g. from demonstration \citep{DBLP:conf/rss/BurkePR19,zhang2018auto}) in a given scenario to generalising that skill (see e.g. \citet{huang2019neural}) by understanding the essential elements and their relations. There are many elements of surfaces, edges, objects in any learning scenario, and the robot will not initially know which are important to the success of the skill. The progression to a generalised skill could involve trial and error while applying to new scenarios, or the application of causal models to explain what happens in a scenario. Here I am only focusing on the projection mechanism that takes an existing model (of the skill here) and applies it to varied scenarios. 
The application constitutes a sort of reasoning or judgement: if the robot finds a mapping of elements and is confident that it should work, then the robot has made the judgement that this scenario is one belonging to the class where the skill can apply.

\subsubsection{Transferring Physics Knowledge}\label{phys}
Sec.~\ref{intro} gave the example of transferring  prior knowledge of moving bodies, e.g., momentum and friction, to a novel scenario (learning to drive). 
In a human this may happen through early development.
Pushing objects on surfaces gives an idea of different frictions, and the experience of overcoming static friction.
The beginnings of momentum may be learned by swinging a heavy object, for example a bucket of water, and discovering the difficulty of stopping it, and its destructive power to move other objects in a collision. 
The combination of friction and momentum ideas can be experienced with  knowledge of running and stopping, leading to sliding on a gravelly surface, or other surface with lower friction.
These are just some example interactions; a human is likely to experience many during development, and may generalise over similar experiences by self-supervised learning \cite[e.g. for prediction, see][]{LeCun2021blog}, producing a generalised `schema' for momentum and another for friction.\footnote{Any newly experienced episode should spawn a schema based on this example of one, and should later become more generalised as further similar experiences are gathered.}
The schema is a model at an abstract level, e.g. for momentum,  representing a generic heavy object and a motion path, and the expectation of difficulty to slow it, or its ability to crash through an obstacle. It is only one step more abstract than  actual experienced episodes; it is not  modern textbook physics knowledge, but would be common to ancient peoples also. It suffices for the knowledge to be implicit and tied to certain experienced contexts, not necessarily generalised to universal rules.
Such a model can then apply by projection to a new scenario with a moving object, to create an appropriate expectation when elements are instantiated (mapped) to that scenario; e.g. to create the expectation in a learner driver of the possibility of sliding on gravel and continuing straight when attempting to make a turn at speed.
Transferring these ideas to robotics requires a similar training regime with diverse experiences, and self-supervised learning to generalise models.
Each model or schema of a physical interaction (like that for momentum above) captures a fragment of physics. A large set of models would approximate human intuitive notions of physics.




\subsection{Projection for High Level Understanding}\label{lang}
Recognising more abstract concepts in videos or text is required for advanced applications in video retrieval, or text understanding (e.g. summarisation). E.g., consider the concepts of betrayal, or falling in love, which might not be explicitly mentioned in the medium at all. 
Recognising such concepts from video or text involve the same reasoning process if both become translated into some abstract representation of scenes and activities happening along a timeline, and higher-level concepts are relations among entities in this abstract description.
Understanding text requires the reader to fit concepts to an ongoing simulation of an unfolding story, entertaining alternatives (where multiple interpretations are possible) and then dismissing alternatives when simulations are inconsistent \citep{Allen:1995:NLU:199291}; understanding video requires the same. 
E.g. imagine that the task is to identify the scene where the first act of betrayal is depicted, and the scene where the betrayal becomes apparent to the betrayed.
We would need a model of betrayal which includes a trust relationship established between two characters at some time, and some discretionary actions taken by one party against the interests of the other at another time, especially when such actions form part of a plan to achieve some goal. Many actions may be ambiguous in isolation, but as plans and goals are gradually recognised those actions take on more definite meanings. This may mean that scenes from an earlier part of the story take on a new interpretation in light of later revelations (i.e. under top-down pressure (projection) from the concept that has been hypothesised to be present). 
Another example is the concept `fraud'. With a sophisticated model of insider fraud in a company, one could develop an AI application that would detect fraud from company accounts by projecting this model. As with the betrayal example, many transactions would be difficult to interpret at first, but with further analysis some of them may fit a model of preparatory behaviour for a fraud attempt. Again we can see the subjective role of the observer in making a judgement about betrayal or fraud, and recall the discussion in Sec.~\ref{robot} which noted that projection is is an act ``through which a cognitive agent asserts its
creative spirit onto the environment'' \citep{INDURKHYA2006133}.

In these examples the reasoning process  involves interaction between bottom-up interpretations and top-down projection from plausible models.
This presumes complex models of plans and concepts such as betrayal. The variety of surface appearances and difficulty of amassing training data make this type of problem more suited to a model projection approach. 
It takes time for humans to learn such models, and a child watching a complex  movie will not interpret the scenes as an adult would, and similarly for a novice inspecting company accounts. 
Even for concepts that children understand, projection is important for them to be able to be recognised in a wide variety of situations. 

Full understanding of natural language may require projection using models learnt outside of language processing, e.g. from sensorimotor interaction. To model a physical situation described in text, and to understand the consequences of various actions or events, one would need at least a partial simulation of the physical situation (see further comments on projection's role in simulation at the end of Sect.~\ref{repr}).
Beyond physical situations, many concepts in language borrow from sensorimotor experiences or other concrete experiences by analogy \citep{lakoff1980metaphors}, e.g. the idea of time `running out' as though it were a container of liquid or particles; human-level understanding of such language may require a link to the physical model, to understand consequences, by analogy, in the same way as a human.
\begin{quote}
    Language does not stand alone. The understanding system in the
brain connects language to representations of objects and situations
and enhances language understanding by exploiting the full
range of our multisensory experience of the world, our representations
of our motor actions, and our memory of previous situations. \citep{McClelland25966}
\end{quote}



\subsubsection{Projection's Role in Novel Analogy}\label{analogy}
The difference between the projection described in the previous subsections and making novel analogies is that above the model being projected is commonly applied to the domain it is projected on, whereas
in a novel analogy the model comes from a domain not usually  applied to this data.
In analogy there is a source and destination, e.g., in the sentence ``The speaker was meandering.'' the source is a river taking a slow twisting path in a valley or floodplain, eroding and depositing sediments; the target is a speaker making a speech.
Novel analogies require  search processes in addition to projection. 
In the meander example: it is necessary to search through aspects of the target `speech' (e.g. the physical movement of the speaker, the pronunciation of words, the choice of topics, etc.) that could potentially serve as elements to be mapped to a model of the source, and it is necessary to search through aspects of the source (e.g. is it the speed, the changing direction, the erosion?) that could be modelled and mapped to the target.\footnote{We are assuming here that the person hearing the sentence is encountering this analogy for the first time. When it becomes familiar no such search is required.}
This is a parallel search process with interaction between preliminary results in the two searches.
Projection is the top-down forcing element in this interaction, because it imposes the structure of the source on the target (the similarity creating aspect noted by \citet{kittay1982creation}). Balancing this in the bottom-up direction is the resistance of the target to having any arbitrary structure imposed on it. In the meander metaphor the final result may vary for different people: some may focus on changing direction applied to topics in the speech, while others may focus on the slow speed and lack of clear purpose.

With metaphorical language a common approach in AI is to paraphrase it into a more common expression which the computer can process with standard techniques \citep{mao2018word}, using dictionary resources which code alternative meanings for common metaphors. This approach loses some nuance of the original meaning. Analogical machinery could help to achieve human-level understanding: the concept in the metaphor would be applied to the elements of the novel domain by projection to create a representation of the analogy in the computer.
In general, language processing requires the fitting of possible candidate concept representations to the ongoing story in a top-down process sharing similarities with model fitting in vision or finding mappings for transfer in robotics. There needs to be considerable flexibility in how concepts can be applied. Consider for example `This can contain the water.' and `We need to contain the fallout from this scandal.' There is a need for a knowledge representation (concept representation) for `contain' or `container' which could give  extreme flexibility in application. 
Even the simplest words seem to require the same kind of projection machinery:
\begin{quote}
To check whether a plate is \textit{on a table} we can look at the
space above the table, but to find out whether a picture is on a wall or a person is on a train, the
equivalent check would fail. A single \textit{on} function operating in the same way on all input domains
could not explain these entirely divergent outcomes of function evaluation. On the other hand, it
seems implausible that our cognitive system encodes the knowledge underpinning these apparently
distinct applications of the \textit{on} relation in entirely independent representations. The findings of this
work argue instead for a different perspective; that a single concept of \textit{on} is indeed exploited in each
of the three cases, but that its meaning and representation is sufficiently abstract to permit flexible
interaction with, and context-dependent adaptation to, each particular domain of application. \citep{DBLP:journals/corr/abs-1902-00120}
\end{quote}

This analogy mechanism can also be applied in robotics to transfer skills across  different domains. 
In the language of mapping  a `source' and a `target', the source is the skill from our repertoire that we want to apply, and the typical situation it has operated in, the target is the new scenario in a different domain that we are attempting to adapt the skill to.
 Sec.~\ref{robot}  assumed that the key elements of the source are known, and only a search of ways to represent the target is required to map components. In a full novel analogy a search of possible ways to represent the source is also required. 
The effect of this type of analogical machinery is that scenes are open to multiple perspectives (as stated in Sec.~\ref{robot}), but also that  each skill can be very general (across domains) as soon as it is learned (although the early applications may not be very successful when the consequences of cross-domain transfer are not yet understood). This may get closer to how humans learn a skill in a context first, and then can quickly generalise that to other contexts.

\section{The Current Status and Prevalence  of Projection in AI}\label{generally}


Despite the advantages that human-like projection could bring to  robotics or advanced vision tasks like video analysis, it is mostly not being researched in these areas; instead the vast majority of work uses feedforward neural networks (FFNNs).
A number of likely reasons include: it is not fashionable at present (many researchers simply follow the dominant trend); people find it easier to improve and extend existing (e.g. FFNN) techniques rather than build a new approach; the  examples of projection in Sec.~\ref{exist} show that considerable effort is needed in designing new models and annotated datasets to capture human-like compositional models \citep{BENYOSEF201865}, as well as iterative inference algorithms \citep{Georgeeaag2612}. This section  considers the status of projection, and related work.

\subsection{Compositional Models}\label{models}

The need for compositional models has been discussed extensively elsewhere \citep{LakeBBSarxiv2016}. A compositional model with relations among parts can accommodate examples that combine parts in ways not seen in training data, and this is a way to overcome the explosion of training data needed for simpler pattern recognition approaches (See \citet{DBLP:journals/corr/abs-1805-04025} Sec. 7 and \citet{franoischollet2017learning} Chap. 9).
For a complex concept, compositional models allow it to  to be recognised in  quite different manifestations, and examples of this were discussed in Sec.~\ref{lang}. Generative models that capture underlying structure allow `analysis by synthesis' or `generative inference' \citep{VANBERGEN2020176}. 
However, although the need for compositional models has been recognised by the prominent researchers cited above\footnote{These are not necessarily widely accepted opinions however, the authors are analysing the limits of the currently popular approaches which are not based on compositional models.}, most current work  does not use compositional models.

Deep learning models learn the manifold of the training distribution, and this limits them to the manifestations in the training distribution, with some interpolations between being successful. If a deep learning model is used to generate samples it can generate unrealistic samples when pushed to generate in regions of the space without training examples, and the way in which its latent space does not capture reality becomes apparent. For example, in GANs generating images conditioned on pose, human legs can be broken in pieces when an out of distribution  pose is forced \citep{ma2017pose}. Efforts to lengthen the sleeve on a T-shirt can lead to the arm taking on an intermediate colour between skin and clothing. The model does not capture the underlying structure of the real world. Humans in contrast can use their background knowledge of underlying real-world structure to go beyond the manifestations they have seen. For example, a human can understand the pose of a ballet dancer even when the body appears in a configuration they have never seen before and are incapable of producing themselves (e.g. with legs making a 180$^{\circ}$ angle while the body stands upright with one leg pointing up and one down). 
To generalise to articulations that are possible but not seen in training examples, it is necessary to model the underlying skeletal structure and the constraints on the joints. It is difficult to see how current methods could learn this from surface appearance without some strong inductive biases.

\subsection{Mapping Models to Data: Inference System}\label{inference}

\citet{Georgeeaag2612} show that iterative inference with a model  is relatively complex, even for simple concepts. \citet[Sec. 4.3.1]{LakeBBSarxiv2016} also discussed the computational challenges of inference with models.  Both of those examples relate to written characters of text. \cite{BENYOSEF201865} do inference with models of small parts of objects. There is a need  to move up to higher level concepts, e.g. human activities, and models of  physical and social interactions.
 We see the outline of the required inference algorithm in \citet{Georgeeaag2612}, and also the iterative inference sketched by \citet{VANBERGEN2020176}: iterations of message passing between higher and lower layers to converge on an interpretation of the data. 
 This paper has focussed on the top down information in `projection' but this is just part of the more general idea of complex iterative inference in which messages may pass between components at various levels of abstraction.
  In general we can expect to have specialised modules to handle parts of interpretation, such tasks as (e.g. for video) object recognition, tracking, atomic actions, assignment of intentions to agents, etc. but no such module should be expected to complete in one pass from input to output, instead they should negotiate with each other to agree an interpretation.
This may require a blackboard or workspace where candidate interpretations are constructed and compete to be the winning interpretation. 
 The workspace would also need to link to world knowledge and simulation to impose plausibility constraints. 
While the effort to build an inference algorithm for complex concepts is daunting, an exciting aspect is that this may  be a `master algorithm' that is used across a wide variety of sensorimotor and cognitive tasks.

\subsection{Models: Representation, Inference, and Learning}\label{model}
Projection relies on hierarchical compositional models. These structures may be hand-designed or learned. In the case that the structures are hand-designed, this is still not like  classical AI which suffered brittleness when outside scenarios the designer envisaged \citep{bro91}. Classical AI assumed a one-to-one mapping between its symbols and entities in the world; the world was to be interpreted in only one way: the `ground truth' interpretation envisaged by the designer.
A system with projection is different because interpretation is not fixed by the designer:  the system itself can search for different ways to map symbols to real-world entities, and can change its interpretation at any time, and can come up with interpretations the designer might not have envisaged.
This means that  projection makes the knowledge structures extremely transferable, to distort the system's perspective on reality and apply to a wide variety of situations (which may entail making mistakes); which is very different to classical AI.

One barrier to implementing projection is the perceived need to learn the models first. It is not clear how this learning should be done in general, for complex concepts, and it is probably difficult. The preference for  learning rather than hand-crafting models may stem from the fear of repeating the mistakes of classical AI with its hand-crated representations that proved brittle when connected to the real world, and also the limited complexity of hand-crafted models. 
However, if work could progress in parallel on the three major AI problems, they might help each other: 1.  Representation (Understanding what models should look like); 2. Inference (Mapping models to data); 3. Learning models.
If we knew more about the types of models that bring classification closer to human level (which \cite{BENYOSEF201865}, e.g., clearly have done), and allow efficient inference, then that would help us to understand how to design learning algorithms that could learn those types of models.
The current dominance of learning in AI acts as a barrier to progress in representation and inference.

\subsection{Related Work in Deep Neural Networks}\label{DNN}
If projection and compositional models are important in human cognition then why are purely feedforward neural networks (FFNNs) fairly successful on some of those tasks that humans can do?
\citet{VANBERGEN2020176} explain how a recurrent neural network (RNN, i.e. with feedback or top-down connections) can be unrolled to a larger FFNN which emulates the RNN, where the recurrent connections become skip (or residual) connections. 
This is an important insight: although we normally think of projection as top-down, the same effect could be achieved in a FFNN. 
Concretely: if a few FFNN layers give us a guess `this might be a horse's head', then we might want to look again at the contours to check if the top of the horses head is indeed there with the shape we expect, hence we can have the result of that lower contour layer passed in to the next FFNN layer again (via skip connection). 
However, the FFNN version of an RNN is less efficient, and some RNNs would lead to infeasible FFNNs if unrolled (ibid.). 
\citeauthor{VANBERGEN2020176} argue for generative models and iterative algorithms for inference, especially to enable ``robustness to variations in the input''. FFNNs need to see examples of variations in training, while generative models can `imagine' combinations not in the training distribution.

Some recent works do bring an element of iterative inference within a neural network framework. \citet{goyal2021coordination} introduce a workspace to which selected specialist modules can contribute information. The contents of the workspace are then broadcast to all modules. Key-query-value attention is used both to decide who gets to write to the space as well as what information modules read from the space afterwards.
In an alternative approach \citet{goyal2021neural} use a working memory to store information about entities and use neural `production rules' to manipulate it. If a rule matches  some entities in memory then it is triggered and a computation is performed to update the working memory. The computation is implemented by a multi-layer perceptron.
 \citet{pmlr-v119-mittal20a} describe experiments with deep recurrent
neural net architectures in which bottom-up
and top-down signals are combined; they use key-query-value attention to be selective about which information is combined.
They applied the system to various tasks and provided insightful analysis of how the top-down information was used. Firstly the top-down information is used very sparingly, e.g. in one experiment only 2.86\% of the total attention was to top-down information. Secondly when noise is added to data there is increased attention to top-down information; ``the model learns to rely
more heavily on expectations and prior knowledge when the
input sequence is less reliable.'' (ibid.)
Thirdly top-down information is relied on more when objects in vision are occluded or partially occluded.
Iterative inference is also a feature of the recent GLOM \citep{hinton2021represent} which aims to represent part-whole hierarchies in visual scenes.

In these works there is a trend towards more iterative inference, which is moving closer to human-like reasoning;
i.e. the trend starting with skip connections, and progressing to attention models which maintain, and iterate with, higher-level hidden states, and then the recent Perceiver \citep{jaegle2021perceiver} where latent units can have multiple interactions with each other and the input data. There is an evolution towards the type of inference which \citet{VANBERGEN2020176} describe.
The following are some major gaps between the above approaches and what needs to be tackled to achieve human-like reasoning:
\begin{itemize}
    \item The tasks tackled and the concepts  are fairly low level, e.g. predicting future frames in a video of MNIST digits moving and bouncing off the frame's edges \citep{pmlr-v119-mittal20a}.
\citeauthor{pmlr-v119-mittal20a} explicitly state that in contrast to other work on higher level entities ``the
current work focuses more on micro modules.'' It is impossible to know if the current approaches are adequate to tackle more advanced concepts until it is attempted, but it seems likely that new approaches would need to be introduced.

    \item Representations of concepts  need to be composed of parts that are similar to the parts humans use, if human-like reasoning is to be emulated. It is not clear if this is necessary in order to have a competent AI system, but large differences in internal representation are likely to lead to different reasoning outcomes in certain situations, e.g. current object recognition systems focus less on contours and more on textures than humans and as a result generalise quite differently to humans \citep{geirhos2019imagenettrained}.
How to discover human-like parts  in the case of vision has been addressed by \citet{BENYOSEF201865}, but less is known about how to determine parts of more abstract concepts.
    \item Iterative inference will need to get more complex than e.g. a single bidirectional interaction with a workspace \citep{goyal2021coordination}. As described by \citet{VANBERGEN2020176}, it may be necessary to move away from the idea of setting the goal as mapping inputs to outputs, but rather to target the network's dynamics; then rather than a fixed number of iterations, the number of interactions could vary, according to the network dynamics, as appropriate to the difficulty of the particular reasoning problem.
    \item There needs to be a way to incorporate background knowledge (all the systems above are  trained from scratch). Humans have background conceptual knowledge that can be applied across varied reasoning tasks. Background knowledge seems to be the only way to deal with high level tasks without needing infeasibly large amounts of training data. 

\end{itemize}




It is difficult to encourage research on more human-like model-based approaches if the performance benefit is not dramatic relative to FFNN approaches. Moving to higher level tasks can be expected to bring a bigger performance difference, where  the FFNN unrolled approach may be infeasible, and where there are more possible diverse manifestations of the same concept, meaning the required training set and training time for an FFNN becomes infeasible.

\subsection{Characteristics of Tasks Suited to Projection}

Projection is not important for every task, this section looks at characteristics of suitable tasks.
Firstly if there are not many examples to train from then the learner will not be able to see the many different ways in which parts can be arranged, and interpolation between examples will be insufficient.  This is exacerbated in situations where the concepts appear in many varied configurations in the data, e.g. the distorted text in CAPTCHAs, or the many ways a betrayal could be depicted in a movie. Projecting a model overcomes this, as described in Sec.~\ref{model}.
\citeauthor{Georgeeaag2612}'s  system boasts much higher accuracy and data efficiency (lower sample complexity) than FFNN approaches.
Secondly if there is high ambiguity in data and the concepts that one is trying to recognise have complex relationships with the data,
then there are three problems:
\begin{enumerate}
\item[1)] 
A bottom-up process can see sufficient evidence for many things in the data, but perhaps only one is correct. This relates to the difficulty of segmenting the characters in recognising CAPTCHAs; a local piece of data could plausibly be a part of a few different concepts (or in understanding movies the piece of data could be a  particular event in a movie). 
\item[2)] 
It is difficult to learn the complex relationships from scratch, from data. When the relationship is only approximated then the model can be fooled by examples which present some strong evidence but not in quite the right configuration. 
\item[3)] 
If there are multiple possible concepts, each requiring different relations, then a FFNN implementation would need to compute all of these in parallel which could be infeasible.  
\end{enumerate}
In contrast the projection  approach can solve problem 1) by a top-down and bottom-up  interaction:
a first bottom-up pass gives preliminary  hypotheses about concepts present (only a subset are really present), and a rendering of those hypotheses through top-down generation can find the minimum number of concepts to explain what part of the data could belong to explained by each concept \citep{Georgeeaag2612}.
This is a type of imagination, and more generally projection should be helpful in cognitive tasks requiring imagination.
Concepts for which all the required evidence is present can still be ruled out if there is a simpler explanation for the data as a whole.
Compositional models can solve problem 2) by first learning sub-concepts, and then learning higher concepts as relations among sub-concepts. Existing FFNNs do this to an extent, but they have not succeeded in learning human-like compositional models, as shown by \citet{BENYOSEF201865} and \citet{BenYosef2018ImageIA}; FFNNs tend to pay more attention to textures and would need some additional biases, and perhaps curriculum learning,  to force them to learn more human-like compositional models.
Problem 3) is solved by only computing relations required by concepts that are likely after a first bottom-up pass \citep{BENYOSEF201865}.  

Many tasks currently successfully tackled by FFNNs do have these characteristics to a degree. Projection becomes more relevant for tasks that have the characteristics more strongly, such as the higher-level tasks outlined in Sect.~\ref{lang}. 
For recognition from still images the following tasks challenge current techniques and projection would be promising for them: abstract recognition\footnote{meaning that the general form of the object is present, but the local features do not resemble those of the real object, for example in recognising a shape resembling an object in clouds. or Fig~\ref{face} (b) and (c).} \citep{dickinson_2009}, when using context is required  to help recognition, when objects occur with heavy occlusion,  or a collage of multiple images \citep{DBLP:journals/corr/RussakovskyDSKSMHKKBBF14} (which is similar to abstract recognition), and also when a small patch needs to be recognised  \citep{Ullman2744} (which is similar to occlusion). 

Finally, projection has a role for general purpose AI:
Most existing AI systems focus on a single task, or fixed set of tasks, such as a robot picking and placing orders, or a system captioning images, or translating language. In the future AI systems will be expected to perform more diverse tasks, until they are expected to have the diversity of humans. Projection then becomes more important, to incorporate top-down input and to create a representation appropriate to the task at hand. In robotics a system would be overloaded if it had to consider every possible manipulation it could potentially do on entering a workshop; instead the robot needs to be directed by its task, and attend to the objects and features that could be used for the task (this is called task-driven representation in Sec.~\ref{robot}). In vision a human looking at a street scene and identifying every object that could be there, and every concept that could be inferred, would likewise be overloaded. Typical computer vision tasks of identifying objects in images or captioning images are quite artificial compared to what humans normally do while carrying out real-world tasks. Humans rarely have the task of identifying everything in an image, or creating a caption for no particular purpose; humans normally approach a scene with a very specific task of finding something they are looking for (not necessarily an object, but perhaps an action or event), this uses top-down direction.

\rd{Conversely, there are also tasks where employing projection will not be beneficial. For example, projection will not be beneficial where ample training data is available to cover all the distribution of cases that will arise at deployment-time. Also in narrow task specific systems, such as the areas where automation is already very successful, projection will not be beneficial.}

\section{Implications for Representation (Concepts, Common Sense, and Situations)}

\label{repr}
Projection provides a way to connect models of knowledge to lower level data, and can bridge the symbolic and sub-symbolic, so it has implications for concepts and commonsense knowledge. 
Connecting commonsense knowledge bases to the world is a major challenge for AI. There have been numerous efforts at creating  knowledge bases designed for commonsense reasoning, however a recent survey of work on benchmarks for commonsense reasoning found that: ``Despite the availability of common and commonsense knowledge
resources discussed in Section 3 [their paper], none of them are actually applied to achieve state-of-the-art performance on the benchmark tasks, and only a few of them are applied in any recent approaches'' \citep[Sec.~4.4]{storks2020recent}; instead they found that people train deep learning models with various datasets, and the knowledge is implicit in the learned model and pre-trained word embeddings.

AI  has for decades attempted to code commonsense concepts, e.g. in knowledge bases, but struggled to generalise the coded concepts to all the situations a human would naturally generalise them to, and struggled to understand the natural and obvious consequences of what it has been told. This led to brittle systems that did not cope well with situations beyond what their designers envisaged. John McCarthy \citeyearpar{McCarthy68programswith} said ``a program has common sense
if it automatically deduces for itself a sufficiently wide class of immediate
consequences of anything it is told and what it already knows''; that is a problem that has still not been solved.
Marvin Minsky \citeyearpar{minskycommonanalogy} estimated that ``Common sense is knowing maybe 30 or 50 million things about the world and having them represented so that when something happens, you can make analogies with others.''	
Minsky presciently noted that common sense would require the  capability to make analogical matches between knowledge and events in the world, and furthermore that a special representation of knowledge would be required to facilitate those analogies.
We can see the importance of analogies for common sense in the way that basic concepts are borrowed, e.g. the tail of an animal, or the tail of a capital `Q', or the tail-end of a temporally extended event (see also examples of `contain', `on', in Sec.~\ref{analogy}).  More than this, for known facts, such as ``a string can pull but not push an object'', an AI system needs to automatically deduce (by analogy) that a cloth, sheet, or ribbon, can behave analogously to the string. For the fact ``a stone can break a window'', the system must deduce that any similarly heavy and hard object is likely to break any similarly fragile material.
Using the language of Sec.~\ref{phys}, each of these known facts needs to be treated as a schema\footnote{Similar facts will be assigned the same schema, which generalises over them.}, and then applied by analogy to new cases.

Projection is a mechanism that can find  analogies (see Sec.~\ref{analogy}) and hence could bridge the gap between models of commonsense concepts (i.e. not the entangled knowledge in word embeddings learnt from language corpora) and text or visual or sensorimotor input.
To facilitate this, concepts should  be represented by hierarchical compositional models, with higher levels describing relations among elements in the lower level components (for reasons discussed in Sec.~\ref{models}). There needs to be an explicit symbolic handle on these subcomponents; i.e. they can not be entangled in a complex network. 
For  visual object recognition a concept can simply be a set of spatial relations among component features, but higher concepts require a complex model involving multiple types of relations, partial physics theories, and causality. 
Secs.~\ref{robot} and \ref{lang} give a hint of what  these concepts may look like, but a full example requires a further paper. 


Moving beyond the recognition of individual concepts, a complete cognitive system needs to represent and simulate what is happening in a situation, based on some input, e.g., text, visual.
This means instantiating concepts in some workspace to flesh out relevant details of a scenario. Sometimes very little data is available for some part of a scenario, and it must be imagined. For example suppose some machine in a wooden casing moves smoothly across a surface, but the viewer cannot see what mechanism is on the underside, the viewer may conjecture it rolls on wheels, and if it gets stuck one may imagine a wheel hitting a small stone. This type of imagination is another projection: assuming a prior model of a wheeled vehicle is available, then the parts of this can be projected to positions in the simulation (parts unseen in the actual scenario). Similarly for a wheel hitting a stone: a schema abstracted from a previously experienced episode of such an occurrence can serve as a model. Simulation and projection must work together to imagine scenarios, because an unfolding simulation may trigger new projections. If the simulation is of something happening in the present, then sensor data can enter to constrain the possibilities for the simulation.

\section{Conclusion}
Projection appears to be a core mechanism working  together with other mechanisms to solve tasks in perception and cognition in humans.
Computational implementations of projection in vision have shown promising results.
It remains to be seen how much benefit the mechanism can bring to other higher-level vision tasks (such as video understanding), and other AI areas such as robotics and language processing. However, as argued above, there are reasons to believe it will be very important in all of these areas.
Finally it may be a key component for solving the longstanding commonsense knowledge problem, because of its ability to map models of knowledge to varied situations, using analogical machinery.
Very little current work in AI attempts to use projection. Deep learning without iterative inference and without using human-like compositional models of concepts is more popular. I hope this paper might encourage more work on projection.



\section*{Acknowledgements}
This paper owes most to Bipin Indurkhya's ideas on projection in human psychology and is essentially exploring their ramifications for AI. Early drafts benefited from extensive comments (and/or discussion, suggested papers) from Yaji Sripada, Chenghua Lin, Ranko Lazic, Lilian Tang, Andrew Gilbert, Quoc Vuong.
The idea of using ``fraud'' as an example concept came from Yaji Sripada. Thanks to Guy Ben-Yosef and Shimon Ullman for further explanation of their top-down process, and  for permission for image use. 
Thanks to the anonymous reviewers for useful comments and references.

\bibliographystyle{apalike}
\bibliography{psych,learning}
\end{document}